\documentclass{llncs}
\usepackage{llncsdoc}

\usepackage{graphicx}
\usepackage[english]{babel}
\usepackage[utf8]{inputenc}
\usepackage{array}
\usepackage{algorithm} 
\usepackage{algpseudocode}
\usepackage{amsmath}

\DeclareMathOperator*{\argmin}{arg\,min}

\begin{document}
\renewcommand{\tablename}{Table}
\floatname{algorithm}{Algorithm}
\renewcommand{\algorithmicrequire}{\textbf{Input:}}
\renewcommand{\algorithmicensure}{\textbf{Output:}}

\title{Fixing Errors  of the Google Voice Recognizer through Phonetic Distance Metrics}
\author{Diego Campos Sobrino\inst{1}, Mario Campos Soberanis\inst{1}, Iván Martínez Chin\inst{1,2}, Víctor Uc Cetina \inst{1,2}}
\institute{SoldAI Research, Calle 22 No. 202-O, García Ginerés, 97070 Mérida, México \\
\email{\{dcampos,mcampos\}@soldai.com} \\
 \and 
Facultad de Matemáticas, Universidad Autónoma de Yucatán, Anillo Periférico Norte, C.P. 97119, Mérida, México\\
\email{imartinezchin@gmail.com,uccetina@correo.uady.mx}}

\maketitle

\begin{abstract}
Speech recognition systems for the Spanish language, such as Google's, produce errors quite frequently when used in applications of a specific domain. These errors mostly occur when recognizing words new to the recognizer's language model or \emph{ad hoc} to the domain. This article presents an algorithm that uses Levenshtein distance on phonemes to reduce the speech recognizer's errors. The preliminary results show that it is possible to correct the recognizer's errors significantly by using this metric and using a dictionary of specific phrases from the domain of the application. Despite being designed for particular domains, the algorithm proposed here is of general application. The phrases that must be recognized can be explicitly defined for each application, without the algorithm having to be modified. It is enough to indicate to the algorithm the set of sentences on which it must work. The algorithm's complexity is $O(tn)$, where $t$ is the number of words in the transcript to be corrected, and $n$ is the number of phrases specific to the domain.\\

\end{abstract}

\section{Introduction}
Traditionally, algorithms used to transform audio to text have been designed using probabilistic models, such as hidden Markov models. However, deep neural networks are also being developed \cite{Becerra2016}, which has made it possible to generate more precise speech recognizers. Nonetheless, when these recognizers are used in particular domains, it is expected that their error will increase. In the context of this article, a particular domain refers to a speech recognition problem where, in addition to recognizing the words of a general language model, it also requires to recognize a set of phrases with Out-of-Vocabulary (OOV) words, created and with meaning exclusively within a particular application. A clear case where these particular domains occur are restaurants, where it is common to put notable names to their dishes or promotions. For example, in Spanish language, the phrase "Jueves Mozzaleroso" (Mozzareloso Thursday), contains "Mozzareloso" as an OOV word used for a particular restaurant context, and it is practically impossible for it to be recognized by a voice recognizer for the Spanish language. To alleviate the problem, Google's recognizer offers the option of specifying a list of phrases from the application domain in order to make them more likely in its language model and thus have a higher probability of being selected as the recognized word. However, this strategy is not as effective in practice, and it is very easy to find examples where it does not work as expected. It is important to emphasize that the algorithm proposed here, despite being designed for particular domains, is of general application. That is, the phrases that must be recognized can be explicitly defined for each application, without the algorithm having to be modified. It is enough to indicate to the algorithm the set of sentences on which it must work.

The article \cite{Harwath2014} proposes using a logistic regression model to classify text correction alternatives in a speech recognition interface, which they claim can reduce the number of possible corrections.

In general, various methods have been proposed in recent years to reduce speech recognizers' errors. A review of the techniques used recently is made in \cite{Errattahi2018}, where the effectiveness of the traditional metrics for evaluating these systems is also questioned. However, the phonetic similarity is not considered for correction nor the evaluation of the results.

Another alternative metric for the evaluation of speech recognition systems proposed in \cite{Favre2013} considers the interpretation of the recognized phrase by a human, which is not necessarily useful when the destination of speech recognition is further language processing by an algorithm.

This article presents an algorithm to correct the errors of the Google speech recognizer for the Spanish language. This algorithm is designed for applications where the words to be recognized are from a particular domain; therefore, the general language model used by the Google recognizer presents errors in its recognition.

The rest of the article is structured as follows. Section 2 formally describes the audio transcription correction problem. In section 3, the algorithm is presented, and its computational complexity is analyzed. Section 4 describes the experimental work carried out with the database of a restaurant application for Spanish language. Finally, in Section 5, the conclusions are provided along with some ideas to develop as future work.

\section{Problem Definition}
Given an audio transcript $T$ of $m$ words, where $n$ words were incorrectly recognized, it is required to correct the recognition errors using an algorithm that is fast enough to be used in real-time.
Commonly four different types of errors can be found in the recognition of words that make up a sentence:

\begin{enumerate}
\item Substitution. When an individual word is incorrectly recognized and substituted for another (e.g., "pistas" instead of "pizzas").
\item Word union. When two or more contiguous words are recognized as one (e.g., "proceso" instead of "por eso").
\item Word separation. When a word is recognized as two or more words in sequence (e.g., "chile ta" instead of "chuleta").
\item Wrong division. When the separation between two words is located in the wrong phoneme (e.g., "pizarra García" instead of "pizza ragazza").
\end{enumerate}

\section{Algorithm} \label{algorithm_section}

The procedure for correcting transcripts proposed in the Algorithm \ref{algo1}, takes as input an audio transcript produced by the speech recognition system (Speech-to-text or STT) and a context composed of a set of phrases of one or more words \textit{ad hoc} to the domain on which the recognition is performed. With those elements, the algorithm verifies the phonetic similarity between the input phrase's recognized words and those provided in the context.  The algorithm transforms the transcript of the STT system and the phrases of the context to their phonetic representation in the IPA system (International Phonetic Alphabet), then analyzes which segments of the transcription are susceptible to be corrected, and calculates their similarity with the Context elements, choosing the contextual phrases with the greatest similarity as candidate phrases. These candidate phrases are considered in descending order by their similarity metric, and if applicable, they replace the corresponding segment of the original transcript.

Let $T_o$ be the audio transcript produced by the speech recognition system and $C = \{c_1, \ldots, c_n \}$ a set of $n$ context-specific phrases, the Algorithm \ref{algo1} modify $T_o$ to produce a corrected transcript $T_c$. To apply this algorithm it is necessary to define the following specifications:

\begin{itemize}
\item A criterion for constructing the $P$ set of words out of context.
\item A window size $v$ of the neighboring region of $p_j$.
\item A construction mechanism of the set $S_j$.
\item An edit distance metric $d(f_s, f_c)$. In this work the Levenshtein distance was used.
\item The decision threshold for the edit distance $u$.
\end{itemize}

\begin{algorithm}
\caption{Transcripts correction algorithm}
\label{algo1}
\begin{algorithmic}[1]
\Require The original audio transcript $ T_o $, $ n $ context-specific phrases $C = \{c_1, \ldots, c_n \}$, a normalized editing distance threshold $u$, and a window size $v$ of neighboring words.
\Ensure A corrected transcript $T_c$.
\item[] \hspace{-0.6cm}\hrulefill
\State Initialize the corrected transcript $T_c = T_o$.
\ForAll{$c \in C$}
	\State Generate the phonetic representation $f_c$ of the phrase $c$.
\EndFor
\State Build the set $P = \{p_1, \ldots, p_m \}$ with the words contained in $T_o$ that are considered out of context.
\ForAll{$p_j \in P$} 
	\State Build a set $S_j$ of sub phrases susceptible to replacement using $v$.
	\ForAll{$s \in S_j$}
		\State Generate the phonetic representation $f_s$ of phrase $s$.
		\State Calculate the normalized edit distance $d(f_s, f_c)$ for all $f_c$.
	\EndFor
	
	\State Select the $(s^*, c^*)$ pair such that $\argmin_{s, c} d(f_s, f_c)$.
	\If{$d(f_s,f_c) < u$}
		\State Add $ (s ^ *, c ^ *) $ to the replacement candidate set $ R $.
	\EndIf
\EndFor
\If {$R \neq \emptyset$}
	\State Sort $R = \{(s^*_1, c^*_1), \ldots, (s^*_L, c^*_L) \}$ ascendingly in $d(f_s, f_c)$.
	\For{$i=1$ to $i=L$}
		\If {$s^*_i$ does not contain words marked in $T_o$}
			\State Replace $ s^*_i$ with $c^*_i$ in $T_c$.
			\State Mark the component words of $s^*_i $ in $T_o$.
		\EndIf
	\EndFor
\EndIf
\State \Return $T_c$
\end{algorithmic}
\end{algorithm}

The complexity of the Algorithm \ref{algo1} is $O(tn)$ and can be calculated as follows. The generation of the phonetic representation in line 3 runs in linear time in the length of the sentence $c$, so we can consider it as a constant $f$. This line, is executed $n$ times since $n$ phrases are considered to exist in $C$. Therefore this routine is performed $fn$ times.

The construction of the set $P$ on line 5 can be done in different ways. If the solution is implemented by comparing all the combinations of the elements of the set $P$ with the elements of the transcription $T_o$, this construction then requires $mt$ operations.

The construction of the set $S_j$ on line 7, for a window $ v = 1 $, requires the creation of 4 sub phrases as follows. Let the pivot word be $p_j$, the set $S_j$ would be formed woth the sub phrases $\{p_j, \hspace{1mm} p_{j-1} p_j, \hspace{1mm} p_jp_{j + 1}, \hspace{ 1mm} p_{j-1} p_jp_{j + 1} \}$. This construction is carried out for each of the $m$ words of $P$. Therefore a total of $4m$ operations is required.

The generation of the phonetic representation of line 9 requires the same number of executions as line 3, that is $f$ repetitions. Since this is repeated for every $m$ word of $P$ and every $4$ elements of ${S_j}$, the total number of times this operation is executed is $4fm$.

Calculating the edit distance on line 10 is done for each combination of the elements of $C$ with the elements of $S_j$, this is $n \times 4$. In turn, this routine is repeated $m$ times since it is in the cycle that begins on line 6. That is, this calculation requires $4mn$ operations.

Selecting the $(s^*, c^*)$ pair of line 12 only requires the same way as line 10, $n \times 4$ comparisons, which are executed for each of the $m$ words in $P$, since it is within the cycle that begins on line 6. That is, this operation is performed $4mn$ times.

The operation of adding $(s^*, c^*)$ to the set $R$ is performed in the worst case $m$ times.

Sorting the pairs $(s_i^*, c_i^*) $ in line 18, can be done in $L = m$ steps at the most, that is $m$ times.

Finally, when $L = m$, the substitution on line 21 and the marking of words on line 22 are performed at most $m$ times each, that is, in $m + m$ operations.

In total, $fn + mt + 4m + 4fm + 4mn + 4mn + m + m + m + m$ operations are required. Where $f$ is considered a constant. That is, the worst-case complexity is determined by calculating the edit distance on line 10 and selecting the pair $(s^*, c^*)$ on line 12, which are $4mn$, that is $O(mn)$, where $m$ is the number of words to be replaced, the worst case being when all the $t$ words of the $T_o$ transcription are replaced, that is , when $m = t$. Therefore we come to the conclusion that the complexity of the Algorithm \ref{algo1} is $O(tn)$.

Now, in a typical dialogue, $T_o$ transcripts rarely exceed 50 words. Also, cases where all the transcript words need to be replaced usually contain less than five words. The above means that the correction of transcripts can be done without any problem in real-time.

\section{Experimental work}

To test the proposed algorithm's correction capacity, an application was implemented on the communication platform \emph{asterisk} that receives telephone calls and captures the acoustic signal of the sentences spoken by the user. Tests were made with users enunciating specific phrases, whose recordings were later sent to the Google service (Google Cloud Speech API) for recognition.

Experiments were carried out using 451 audio files recorded by nine different users via telephone within the context of picking up orders from a pizzeria. The examples were simulated based on conversations of lifting orders from a real pizzeria's menu. As usual, the menu has different ingredients, specialties, and packages, many of which contain words from languages other than Spanish or invented names with characteristics that make them difficult to identify for speech recognition systems with models of general-purpose language.

Each example was recorded in a \emph{flac} \cite{Flac} format file, with the actual phrase uttered by the user and the transcripts obtained as a result of sending the audio signal to Google stored in a database. We used the Google service in two different modalities, the first in the basic mode and the second, including as context, 34 phrases typical of the pizzeria domain, which according to the Google API documentation, are favored during the recognition process, thus improving the accuracy of the results.

The requests to the speech recognition API were made by sending an HTTP request through the POST method to the URL https://speech.googleapis.com/v1/speech: recognize with the following configuration in \emph{JSON} format:

\begin{verbatim}
{
    config: {
    	encoding: 'FLAC',
    	sampleRateHertz: 8000,
    	languageCode: 'es-US',
    	profanityFilter: false,
    	maxAlternatives: 1
    },
    audio: {
    	content: base64String
   	}
}
\end{verbatim}

In the context recognition mode, the \emph{SpeechContext} property was added to the request, as an array containing the context-specific phrases. A set of 34 phrases typical of the pizzeria domain, between one and three words in length, was used. Phrases may contain variations that are misspelled but phonetically similar to the correct pronunciation. The \emph{JSON} object used has the following structure:

\begin{verbatim}
{
    config: {
        encoding: 'FLAC',
        sampleRateHertz: 8000,
        languageCode: 'es-US',
        profanityFilter: false,
        maxAlternatives: 1
    },
    audio: {
        content: base64String
    },
    SpeechContext: [
        barbiquiu,
        buchelati,
        bustarela,
        dipdish,
        extra pepperoni,
        jueves mozzareloso,
        pizza de corazón,
        pizza ragazza,
        ...
    ]
}
\end{verbatim}

The transcripts obtained from the speech recognition service were processed using the phonetic correction algorithm described in section \ref{algorithm_section}. As context $C$, the same set of 34 phrases provided to the Google service in its contextual mode was used.

The algorithm specifications were as follows:
\begin {itemize}
\item The set $ P $ corresponds to all the words present in the input transcript that are not found in $ C $ and with a minimum size of 4 characters. $P = \{p \hspace{1mm} | \hspace{1mm} p \in T_o, \hspace{1mm} p \notin C, \hspace{1mm} len(p) \geq 4 \} $
\item The window size was $v = 1$.
\item For each word $p_j$ its immediate neighbors were considered, constructing the set as follows $ S_j = \{p_{j}, \hspace {1mm} p_{(j-1)} p_{j}, \hspace{1mm} p_{j} p{(j + 1)}, \hspace{1mm} p{(j-1)} p_{j} p{(j + 1)} \} $.
\item The function $ d(f_s, f_c)$ used was the standard Levenshtein distance \cite{Levenshtein1966} with unit cost for deletions, insertions and substitutions. This metric was normalized in relation to the maximum size of the input sentences.
\item We experimented with different decision thresholds $u$ for the function $ d(f_s, f_c)$ to observe the impact of this parameter on the results of the algorithm.
\end{itemize}

Various methods have been proposed to compare the performance of speech recognition systems \cite{Gillick1989} \cite{Pallett1990}, the most common being the use of the WER (\textit{word error rate}) metric. In this work, the proposed algorithm's results were evaluated using said metric between the target phrase and the recognized hypothetical transcript. WER is defined as follows:
\begin{equation}
WER = \dfrac {S + D + I} {N}
\end{equation}

where $S$ is the number of substitutions, $D$ the number of deletions, $I$ the number of insertions, and $N$ the number of words in the target phrase. The $S$, $D$, $I$, and $N$ values were globally accumulated, calculating the number of edits required to transform the hypothetical transcript into the correct target phrase for each one of the 451 examples.

When calculating WER for both versions of the Google service results, two baselines were obtained that serve as a reference to evaluate the results of the proposed algorithm. The WER obtained with the basic service transcript was $33.7\% $, while the transcript with the contextualized service had a WER of $31.1\%$. Although these values seem high in relation to the results reported in the recognition systems of general use, it must be considered that the set of example sentences is very domain-specific, which increases the difficulty of the problem. Besides, the degradation in the signal produced by the use of conventional telephone lines also affects performance.

\begin{figure}
	\centering
	\includegraphics[width=0.9\textwidth]{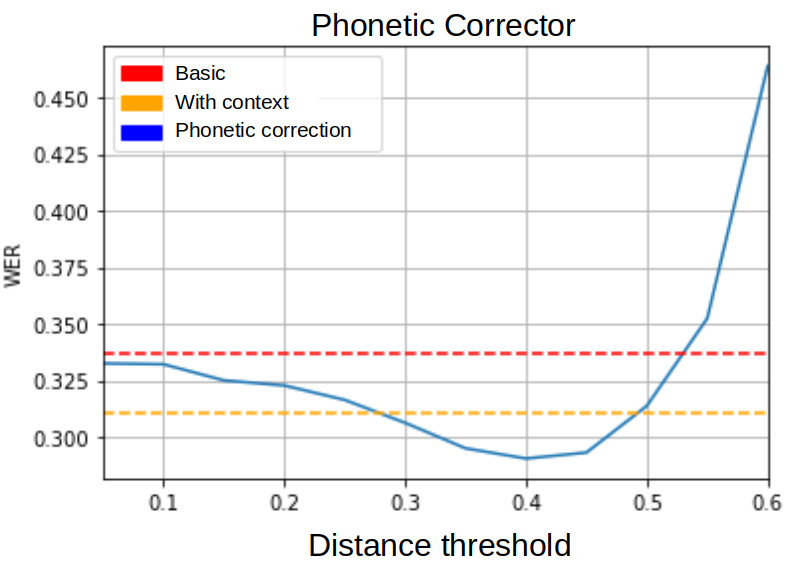}
	\caption{Results of the algorithm for different values of $ u $ taking as input the basic transcript of Google's TTS service.}
	\label{fig:comparison_basic}
\end{figure}

Figure \ref{fig:comparison_basic} shows the algorithm's behavior when varying the $u$ parameter when it is executed taking as input the transcripts obtained with the basic service. The best result is obtained when $u = 0.4$, where the WER obtained is reduced to $29.1\%$. With the basic transcript as a baseline, the result improves global WER by $4.6\%$. The number of example sentences improved in their recognition was 97 out of 325 sentences that were erroneously recognized by the Google service.

\begin{figure}
	\centering
	\includegraphics[width=0.9\textwidth]{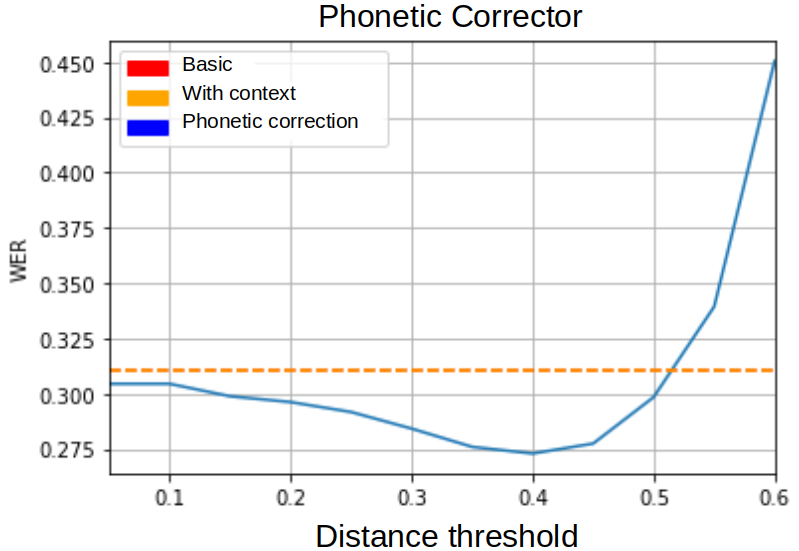}
	\caption{Results of the algorithm for different values of $ u $ taking as input the transcription with context of Google TTS service.}
	\label{fig:comparison_context}
\end{figure}

Figure \ref{fig:comparison_context} presents the results when the context phrases are submitted to Google service, and the phonetic correction algorithm subsequently processes the transcripts. In this case, an improvement in WER is observed, reaching a minimum of $ 27.3\%$ also with the parameter $u = 0.4$. With this value, WER improves by $3.8 \%$ compared to contextual transcription and $6.4\%$ compared to basic transcript. In this case, the number of improved sentences was 87 out of a total of 319 wrongly recognized by Google.

Table \ref{tab:wers} indicates the number of recognition errors for the total of 2664 words contained in the audio examples. From basic recognition, the phonetic correction algorithm produces a reduction in the relative WER of 12.7\%. When the algorithm is applied to the contextual version of the recognizer, the relative improvement in WER is 10.3\%.

\begin{table}
	\center
	\begin{tabular}{| c | c | c | c |}
		\hline
		\textbf{Recognition mode} & \textbf{TTS errors} & \textbf{Corrector errors} & \textbf{Relative WER} \\
		\hline
		Basic & 897 & 774 & 13.6 \% \\
		\hline
		With context & 828 & 727 & 12.2\% \\
		\hline		
	\end{tabular}
	\caption{Number of edition errors and relative WER.}
	\label{tab:wers}
\end{table}

\begin{figure}
	\centering
	\includegraphics[width=0.9\textwidth]{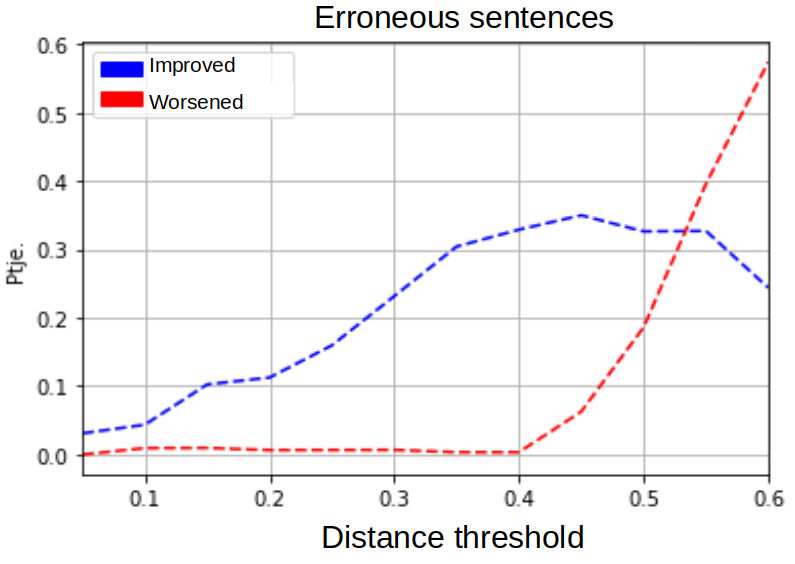}
	\caption{Percentage of improved sentences in relation to the total of sentences with recognition errors in the basic transcript.}
	\label{fig:sentences_basic}
\end{figure}

\begin{figure}
	\centering
	\includegraphics[width=0.9\textwidth]{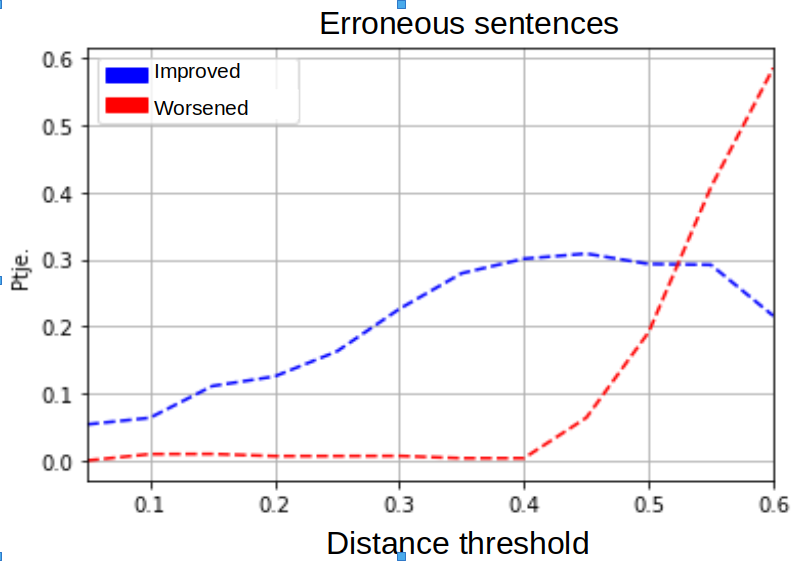}
	\caption{Percentage of sentences improved in relation to the total of sentences with recognition errors in the transcript with context.}
	\label{fig:sentences_context}
\end{figure}

Figures \ref{fig:sentences_basic} and \ref{fig:sentences_context} shows the behavior in the percentage of erroneous sentences corrected by varying the distance threshold from the transcripts in the two speech recognition modalities. It is observed that in both cases when the value of $u = 0.4$ is exceeded, phonetic corrections begin to worsen recognition instead of improving it.

Some examples where the correction process improves recognition are seen in Table \ref{tab:phrases}. Cases are shown where the algorithm manages to correct the phrase recognized by Google completely. In contrast, in other cases where the transcript show a high WER, it is possible to improve enough to identify the phrase's context.

\begin{table}
	\center
	\begin{tabular}{|>{\hspace{1pc}} l <{\hspace{1pc}}|>{\hspace{1pc}} l <{\hspace{1pc}}|}
		\hline
		\textbf{G. P.} & Mándame una Buscar ella\\
		\textbf{C. P.} & Mándame una bustarella\\
		\textbf{T. P.} & Mándame una bustarella\\
		\hline \hline
		\textbf{G. P.} & Voy a querer una grande de chile ta\\
		\textbf{C. P.} & Voy a querer una grande de chuleta\\
		\textbf{T. p.} & Voy a querer una grande de chuleta\\
		\hline \hline
		\textbf{G. P.} & 2 pizzas medianas y clover\\
		\textbf{C. P.} & 2 pizzas medianas meat lover\\
		\textbf{T. P.} & 2 pizzas medianas meat lover\\
		\hline \hline
		\textbf{G. P.} & La pizarra García mediana\\
		\textbf{C. P.} & La pizza ragazza mediana\\
		\textbf{T. P.} & Una pizza ragazza mediana\\
		\hline \hline
		\textbf{G. P.} & Pistas de Barbie dress up\\
		\textbf{C. P.} & Pizzas de barbecue dress up\\
		\textbf{T. P.} & Dos pizzas de barbecue con mucho queso\\
		\hline \hline
		\textbf{G. P.} & Quiero un vitel aquí\\
		\textbf{C. P.} & Quiero un Buccellati\\
		\textbf{T. P.} & Quiero un Buccellati\\
		\hline \hline
		\textbf{G. P.} & Un paquete de jugadores mozzareloso\\
		\textbf{C. P.} & Un paquete de jueves mozzareloso\\
		\textbf{T. P.} & Un paquete de jueves mozzareloso\\
		\hline
	\end{tabular}

	\caption{Examples of phrases enhanced by the phonetic correction process. For each example, the Google Phrase (G. P.), the Corrected Phrase (C. P.) phonetically, and finally the Target Phrase (T. P.) are provided.}
	\label{tab:phrases}
\end{table}

\section{Conclusions and future work}
In this article, an algorithm has been proposed to correct Google speech recognizer errors in domain-specific applications. In these scenarios, the advantage is that it is possible to generate a reduced dictionary of words \emph{ad hoc} to the application, and this dictionary can be used to correct recognition errors. In the proposed algorithm, the Levenshtein distance on phonemes is used to assign indices to the candidate words to be used in the corrections. The algorithm was experimentally tested for the Spanish language, but it is general enough to be used with other languages. As a case study, phrases from an automated system that takes take-out orders were used in pizza restaurants. The experimental results show that using this algorithm, the recognizer errors can be reduced by up to $5.8\%$ compared to Google service's basic transcript. The algorithm manages to improve the recognition in around $30\%$ of the phrases that contain errors. Although not in all cases the correction is perfect; it is possible to improve enough to understand the context of the sentence, which is more important considering that said corrected transcription is frequently used as input for intent classification and entity recognition algorithms, which are characteristic of an automated natural language understanding system.

During experimentation, there were some cases where the algorithm produces an erroneous artifact; as an example, for the input transcript "En que consiste el jueves mozart el oso" the result is "En que consiste el jueves mozzareloso oso". It is possible that these types of cases can be corrected by different selection criteria of the $S_j$ sets in the algorithm, which undoubtedly requires more careful analysis. Future work is also envisioned to explore different editing distances, including the costs for different types of editing, such as deletions, insertions, substitutions, and transpositions.

\end{document}